\documentclass[runningheads]{llncs}
\usepackage{epsfig}
\usepackage{subfigure}
\usepackage{graphicx}
\usepackage{epstopdf}
\usepackage{float}
\usepackage{color}
\usepackage{cite}
\usepackage{multirow}
\usepackage[colorlinks,linkcolor=black,citecolor=black,anchorcolor=black,urlcolor=blue]{hyperref}

\newcommand{\Chen}[1]{\textcolor{red}{#1}}
\newcommand{\lxp}[1]{\textcolor{blue}{#1}}

\begin{document}
\title{{3D Dilated Multi-Fiber Network for Real-time Brain Tumor Segmentation in MRI}}

\titlerunning{3D DMFNet for real-time brain tumor segmentation}

\author{Chen Chen$^{1,}$\thanks{Chen Chen and Xiaopeng Liu contributed equally.},
Xiaopeng Liu$^{2,\star}$,
Meng Ding$^3$, Junfeng Zheng$^2$, Jiangyun Li$^{2,\dagger}$}

\institute{
Department of Electrical and Computer Engineering, University of North Carolina at Charlotte, USA, \email{chen.chen@uncc.edu}\\
\and
School of Automation and Electrical Engineering, University of Science and Technology Beijing, China,
\email{s20170594@xs.ustb.edu.cn, g20188721@xs.ustb.edu.cn, leejy@ustb.edu.cn}\\
\and
Scoop Medical,
\email{meng.ding@okstate.edu}\\
$\dagger$ Corresponding author: Jiangyun Li
}

\maketitle
\begin{abstract}
Brain tumor segmentation plays a pivotal role in medical image processing. In this work, we aim to segment brain MRI volumes. 3D convolution neural networks (CNN) such as 3D U-Net~\cite{cciccek20163d} and V-Net~\cite{vnet} employing 3D convolutions to capture the correlation between adjacent slices have achieved impressive segmentation results. However, these 3D CNN architectures come with high computational overheads due to multiple layers of 3D convolutions, which may make these models prohibitive for practical large-scale applications. To this end, we propose a highly efficient 3D CNN to achieve real-time dense volumetric segmentation. The network leverages the 3D multi-fiber unit which consists of an ensemble of lightweight 3D convolutional networks to significantly reduce the computational cost. Moreover, 3D dilated convolutions are used to build multi-scale feature representation. Extensive experimental results on the BraTS-2018 challenge dataset show that the proposed architecture greatly reduces computation cost while maintaining high accuracy for brain tumor segmentation. The source code is available at \url{https://github.com/China-LiuXiaopeng/BraTS-DMFNet}

\end{abstract}

\section{Introduction}

Recent advances in the treatment of gliomas have increased the demands on using magnetic resonance imaging (MRI) techniques for the diagnosis, tumor monitoring, and patient outcome prediction. Accurate segmentation of brain tumor is critical for diagnosis and treatment planning. However, automated brain tumor segmentation in multi-modal MRI scans is a challenging task due to the heterogeneous appearance and shape of gliomas~\cite{bakas2018}.

A flurry of research has leveraged Convolution Neural Networks (CNNs) for brain tumor segmentation and achieved great success. Havaei \textit{et al.}~\cite{havaei2017brain} present a two-pathway CNN architecture and predict the label for each pixel by taking as input a local image patch in a sliding-window fashion. Ronneberger \textit{et al.}~\cite{ronneberger2015u} develop a fully convolutional network (FCN), namely U-Net, to process the entire image for dense prediction. The network follows an encoder-decoder structure and is trained end-to-end to produce a full-resolution segmentation. Although these 2D CNN-based approaches have achieved impressive performance, these models ignore crucial 3D spatial context given that most clinical imaging data are volumetric, e.g. 3D MR images. To better represent the 3D volumes of imaging data,  Cicek \textit{et al.} \cite{cciccek20163d} generalize the U-Net from 2D to 3D by exploring 3D operations, e.g. 3D convolution and 3D max pooling, in the FCN, leading to the 3D U-Net. Similarly, V-Net \cite{vnet} uses volumetric convolutions to process MRI volumes and yields more accurate segmentation than the 2D approaches.


It has been shown that using 3D convolutions in deep neural networks is an effective way of reasoning volumetric structure~\cite{cciccek20163d,vnet,dou20173d}. However, using multiple layers of 3D convolutions suffers from high computational cost compared with regular 2D CNNs due to an extra dimension. A few attempts have been made to alleviate this issue by using light-weight network architectures. For example, 3D-ESPNet \cite{nuechterlein20183d} extends ESPNet, a fast and efficient network based on point-wise convolution for 2D semantic segmentation, to 3D medical image data. SD-UNet \cite{chen2018s3d} takes advantages of the separable 3D convolution, which divides each 3D convolution into three parallel branches, in order to reduce the number of learnable network parameters.
However, the performance of these efficient models is not comparable to the state-of-the-art.

\textbf{Contribution.} In this paper, to bridge the gap between model efficiency and accuracy for 3D MRI brain tumor segmentation, we propose a novel 3D dilated multi-fiber network (DMFNet). It builds upon the multi-fiber unit \cite{chen2018multi}, which uses the efficient group convolution, and introduces a weighted 3D dilated convolution operation to gain multi-scale image representation for segmentation. Note that Qin \textit{et al.} \cite{qin2018autofocus} propose the auto-focus convolution layer to improve brain tumor segmentation via the adaptive weighted dilated convolution. The main differences lie in that they impose strong constraints on the dilated convolution, i.e. using softmax to constrain the weights of the parallel dilated convolution branches and sharing the parameters of the convolution kernels.
Our DMFNet only has 3.88M parameters.
With the inference times of 0.019s on one GPU and 20.6s on one CPU for a single 3D volumetric segmentation, it achieves dice scores of 80.12\%, 90.62\% and 84.54\% respectively for the enhancing tumor, the whole tumor and the tumor core on the 2018
BraTS challenge~\cite{bakas2017advancing,menze2015multimodal}.

\section{Method}
\subsection{Dilated Multi-Fiber (DMF) Unit}

3D convolution kernel is normally operated on the entire channels of the feature maps, which scales up the computational complexity exponentially in terms of floating point operations per second (FLOPs).
Group convolution is an effective solution for model speeding up, which has been explored for efficient network design, e.g. ShuffleNet \cite{shufflenet}. Although the grouping strategy could reduce the number of parameters, simply replacing the regular convolution with the group convolution may impact the information exchange between channels and hurt the learning capacity. Multi-fiber (MF) \cite{chen2018multi} is proposed for video action recognition and can facilitate information flow between groups. Inspired by that, we extend the multi-fiber unit design with an adaptive weighted dilated convolution to capture the multi-scale features in brain MR images.
In the following, we detail the key components of our DMF unit.


\begin{figure}[!t]
  \centering
  \includegraphics[width=\textwidth]{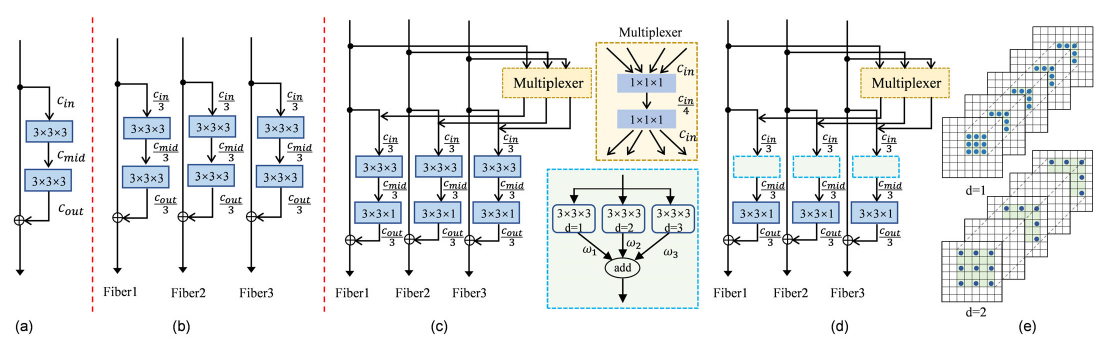}
  \caption{(a) A residual unit with two regular convolution layers. (b) The multi-fiber design consisting of multiple separated residual units, called fibers. (c) The multi-fiber (MF) unit takes advantage of a \textit{multiplexer} for information routing. (d) The proposed dilated multi-fiber (DMF) unit with an adaptive weighting scheme for different dilation rates. (e) The schematic diagram of the 3D dilated convolution operation. $d$ is the dilation rate. $d=1$ indicates the regular convolution.}
  \label{MFunit}
\end{figure}

\par\textbf{Channel Grouping.} The idea of channel grouping is to split the convolutional channels as multiple groups that can reduce the connections between the feature maps and kernels for parameter saving significantly. As examples shown in \mbox{Fig. \ref{MFunit}} (a) and (b), the regular residual unit is grouped into $g$ parallel residual units that are called fibers. We assume the kernel size is constant, e.g. $kernel=3\times 3\times 3$ and denote $param_{(a)}$ and $param_{(b)}$ as the parameter amounts of \mbox{Fig. \ref{MFunit}} (a) and (b), respectively. Thus, we have $param_{(a)}=kernel\times(c_{in}\times c_{mid}+c_{mid}\times c_{out})$, where ${c_*}$ is the number of channel. With the strategy of multiple fibers grouping, the amount of parameter comes to $param_{(b)}=g\times kernel\times (c_{in}/{g}\times{c_{mid}}/{g}+{c_{mid}}/{g}\times{c_{out}}/{g})=param_{(a)}/{g}$, which is $g$ times less than $param_{(a)}$.

\par\textbf{Multiplexer.} To facilitate the information exchange between fibers, the $1\times 1\times 1$ convolutions, dubbed as multiplexer, are utilized for information routing among different fibers \cite{chen2018multi}. It is comprised of two $1\times 1\times 1$ convolution layers, as illustrated in \mbox{Fig. \ref{MFunit}}. And the input channel $c_{in}$ is squeezed to $c_{in}/4$ and then inflated to $c_{in}$.
By employing two $1\times 1\times 1$ convolutions ($params=c_{in}\times c_{in}/4+c_{in}/4\times c_{in}=c_{in}^2/2$), it can reduce half of the parameters as compared to using one $1\times 1\times 1$ convolution ($params=c_{in}^2$). {Besides, the residual shortcuts,  which are placed outside the multiplexer and the entire unit, allow the information pass through from lower level to higher level directly, leading to enhanced learning capability without additional parameters.}

\par\textbf{Dilated Fiber.} To enlarge the respective field and capture the multi-scale 3D spatial correlations of the brain tumor lesions, the dilated convolution~\cite{dilated} is employed. As shown in \mbox{Fig. \ref{MFunit} (d)}, the dilated fiber is comprised of three 3D dilated convolution branches with the dilation rates of $d=1$, 2 and 3 respectively. We allocate the learnable weights $\omega_1$, $\omega_2$ and $\omega_3$ to each dilated branches, and then sum them up. This weighted sum strategy is conductive to select most valuable information automatically from different field of view. The weight coefficients are one-initialized, which means the branches contribute equally at the beginning of the training process.



\subsection{Dilated Multi-Fiber Network (DMFNet) Architecture}

\begin{figure}[htbp]
	\includegraphics[width=\textwidth]{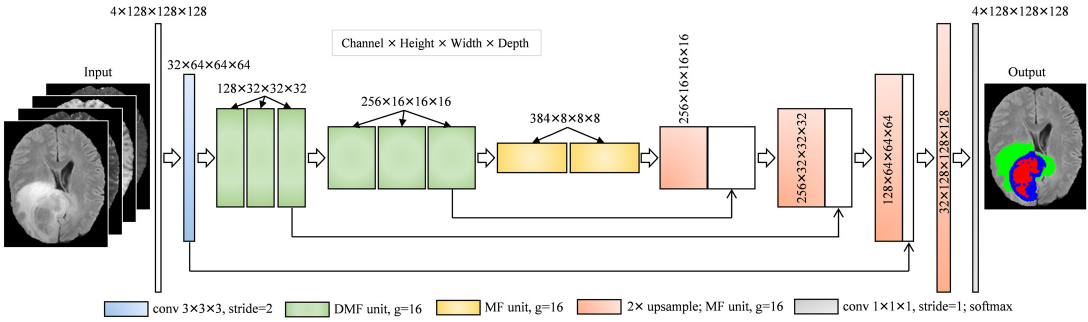}
	\caption{The proposed dilated multi-fiber network for 3D MRI brain tumor segmentation, where $g$ is referred to the number of groups, e.g. $g=16$ used in this work. We use 2-stride convolution to downsample the feature maps.}
	\label{Architecture}
\end{figure}

Using the MF and DMF units as the building blocks, the overall encoder-decoder network architecture of DMFNet is shown in \mbox{Fig. \ref{Architecture}}. The 4-channel input corresponds to 4-modal MRI data. The main body of the network is composed by the MF/DMF units, excluding the first and last convolution layers. In the feature encoding stage, we apply the DMF unit in the first six encoding units to achieve multi-scale representation, which is benefited by the various sizes of receptive field in the dilated convolution. In the decoding stage, the high resolution features from the encoder are concatenated with the upsampled features, which is similar to the U-Net. We adopt the trilinear interpolation for upsampling the feature maps. Also, batch normalization and ReLU function are performed before each convolution operation of MF/DMF units.

\section{Experiments and Results}
\subsection{Experimental setup}
\noindent \textbf{Data and evaluation metric.} The 3D MRI data, which is provided by the Brain Tumor Segmentation (BraTS) 2018 challenge~\cite{menze2015multimodal,bakas2017advancing}, consists of four kinds of MR sequences, namely native T1-weighted (T1), post-contrast T1-weighted (T1ce), T2-weighted (T2) and Fluid Attenuated Inversion Recovery (FLAIR). Each of them has a volume of $240\times240\times155$. The labels for tumor segmentation include the background (label 0), necrotic and non-enhancing tumor (label 1), peritumoral edema (label 2) and GD-enhancing tumor (label 4).
The dataset consists of 285 cases of patients for training and 66 cases for validation.
The performance on the \textbf{validation set} assessed by the online evaluation server is used to validate the effectiveness of the proposed method. Specifically, the effectiveness is evaluated by the computational complexity and the segmentation accuracy. The complexity is determined by the amount of network parameters and FLOPs (i.e. multiplication and addition)~\cite{shufflenet}. The segmentation accuracy is measured by the dice score metrics and Hausdorff95 distance metric, where ET, WT and TC are referred to the regions of enhancing tumor (label 1), the whole tumor (label 1, 2 and 4) and the tumor core (label 1 and 4) respectively.


\textbf{Implementation details.} In our experiments, we use a batch size of 12 and train the DMFNet model on 4 parallel Nvidia GeForce 1080Ti GPUs for 500 epochs.
We adopt the Adam optimizer with an initial learning rate ${\alpha}_0=0.001$.
To increase the training data, we use the following data augmentation techniques: (\romannumeral1) random cropping the MRI data from $240\times 240\times 155$ voxels to $128\times 128\times 128$ voxels; (\romannumeral2) random mirror flipping across the axial, coronal and sagittal planes by a probability of 0.5; (\romannumeral3) random rotation with the angle between $[-10^{\circ},+10^{\circ}]$; (\romannumeral4) random intensity shift between $[-0.1,0.1]$ and scale between $[0.9,1.1]$.
The generalized dice loss (GDL) is employed to train the network. L2 norm is applied for model regularization with a weight decay rate of $10^{-5}$. In the testing phase, we zero pad the $240\times 240\times 155$ MRI data in the depth dimension to $240\times 240\times 160$ (depth dividable by 16) as the network input.

\subsection{Experimental results and analysis}

\begin{figure}[!t]
    \centering
    \includegraphics[width=10cm]{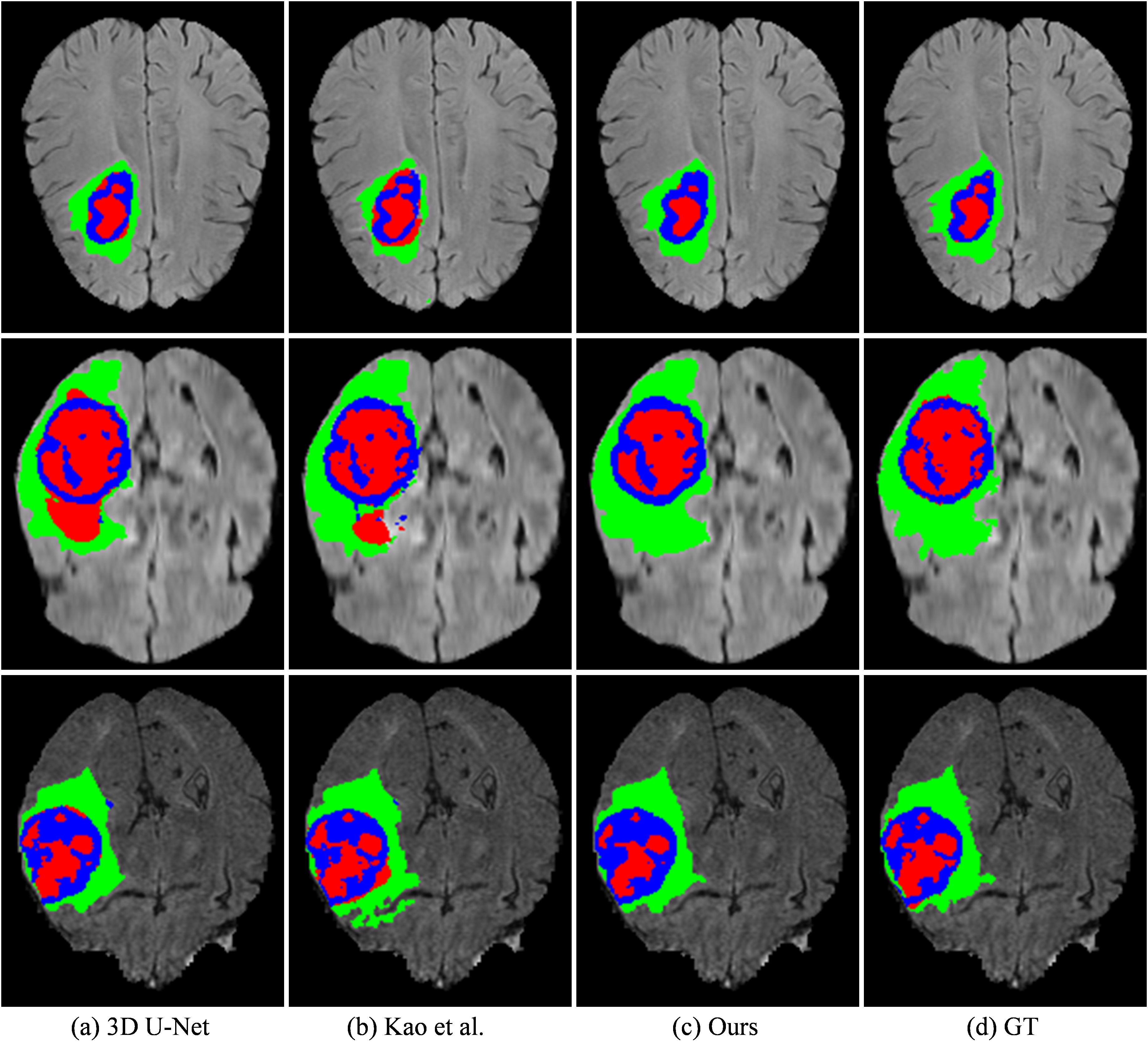}
    \caption{The visual comparison of MRI brain tumor segmentation results. GT indicates the ground-truth. The regions in red represent the necrotic and non-enhancing tumor, the regions in green represent the peritumoral edema and the regions in blue represent the GD-enhancing tumor.}
    \label{VisFig}
    \end{figure}

    \textbf{Comparison with state-of-the-art.}  We first conduct five-fold cross-validation evaluation on the training set, our DMF-Net achieves average dice scores of 76.35\%, 89.09\% and 82.7\% respectively for the enhancing tumor, the whole tumor and the tumor core. We also carry out experiments on the BraTS 2018 validation set and compare our method with the state-of-the-art approaches. The performance comparison is presented in \mbox{Table \ref{table2}}. Our proposed DMFNet achieves scores of 80.12\%, 90.62\% and 84.54\% for ET, WT and TC, respectively. Compared to the best scores achieved by NVDLMED~\cite{myronenko20183d} (single model), it can be seen that our model only has marginal performance gaps of $0.06\%$ for the whole tumor, $1.61\%$ for the enhancing tumor and $1.48\%$ for the tumor core respectively. However, our DMFNet has $10\times$ less parameters and $55\times$ less FLOPs. Therefore, our method is a much more efficient algorithm yet can achieve comparable segmentation accuracy. We also show a visual comparison of the brain tumor segmentation results of various methods including 3D\_UNet \cite{cciccek20163d}, Kao \textit{et al.} \cite{kao2018brain} and our DMFNet in \mbox{Fig. \ref{VisFig}}. It is obvious that DMFNet is able to generate better segmentation (especially at the class boundaries) due to the multi-scale representation of dilated convolutions.

\begin{table}
\caption{Performance comparison on the BraTS 2018 validation set.}
\centering
\begin{tabular}{l|cc|ccc|ccc}
\hline
\label{table2}
\multirow{2}{*}{\textbf{Model}} &\multirow{2}{*}{\textbf{Params(M)}} &\multirow{2}{*}{\textbf{FLOPs}}
&\multicolumn{3}{c}{\textbf{Dice\_score(\%)}} &\multicolumn{3}{c}{\textbf{Hausdorff95}} \\ \cline{4-9}
& & & ET & WT & TC & ET & WT & TC \\
\hline
0.75$\times$ MFNet (\textbf{ours}) & \textbf{1.81} & \textbf{13.36} & 79.34 & 90.22 & 84.25 & {2.59} & {4.60} & {5.87} \\
MFNet (\textbf{ours}) & {3.19} & {20.61} & {79.91} & {90.43} & {84.61} & {2.68} & {4.68} & {6.31} \\
DMFNet (\textbf{ours}) & 3.88 & 27.04 & 80.12 & 90.62 & 84.54  & {3.06} & {4.66} & {6.44}\\
\hline
3D U-Net~\cite{cciccek20163d} & 16.21 & 1669.53 & 75.96 & 88.53 & 71.77  & {6.04} & {17.10} & {11.62}\\
S3D-UNet~\cite{chen2018s3d} & 3.32 & 75.20 & 74.93 & 89.35 & 83.09  & {-} & {-} & {-}\\
3D-ESPNet~\cite{nuechterlein20183d} & 3.63 &  76.51 & 73.70 & 88.30 & 81.40 & {-} & {-} & {-} \\
Kao et al.~\cite{kao2018brain} & 9.45 & 203.96 & 78.75 & 90.47 & 81.35 & {3.81} & {4.32} & {7.56}\\
No New-Net~\cite{isensee2018no} & 10.36 & 202.25 & 81.01 & \textbf{90.83} & 85.44 & {2.41} & {4.27} & {6.52}\\
NVDLMED~\cite{myronenko20183d} & 40.06 & 1495.53 & \textbf{81.73} & 90.68 & \textbf{86.02} & {3.82} & {4.52} & {6.85} \\

\hline
\end{tabular}
\end{table}

    \textbf{Model efficiency.} It is also evident from \mbox{Table \ref{table2}} that our DMFNet significantly outperforms the methods which have similar or close model complexity (\# of parameters and FLOPs), i.e. S3D-UNet and 3D-ESPNet. Without using the dilated convolution, the 3D MFNet further reduces the model complexity. Moreover, we devise a remarkably lightweight and efficient network (denoted by 0.75$\times$ MFNet in \mbox{Table \ref{table2}}) by reducing the number of channels in MFNet (see \mbox{Fig. \ref{Architecture}}) to 75\%. Therefore, it has only 1.81M parameters and 13.36G FLOPs. Nevertheless, its dice scores still reveal the network has strong learning capability for 3D brain tumor segmentation. In addition, DMFNet obtains an average inference time of 0.019s on one GPU (Nvidia 1080Ti) or 20.6s on one CPU (E5-2690 v3 @ 2.60GHz) for a single 3D MR image segmentation.

    \textbf{Ablation study.} The performance comparison between MFNet and DMFNet (\mbox{Table \ref{table2}}) demonstrates that the dilated convolution is able to boost the dice scores. Since an adaptive weighting strategy is used for convolutions with different dilation rates (Fig. \ref{MFunit} (d)), its efficacy is justified in \mbox{Table \ref{table4}} by comparing it with the equal weight scheme ($\omega_1=\omega_2=\omega_3=1$). Due to the ability of learning and selecting the multi-scale context information adaptively, such weighting strategy results in more favorable scores, in particular for the ET score.

    \begin{table}
	\centering
	\caption{The effect of the weighting strategy of the dilated sub-fibers.}
	\label{table4}
    \begin{tabular}{c|c|ccc}
    	\hline
            	\multirow{2}{*}{\textbf{Model}} &\multirow{2}{*}{\textbf{Weighting scheme}} &\multicolumn{3}{c}{\textbf{Dice\_score(\%)}}  \\ \cline{3-5} & & ET & WT & TC \\

        \hline
        DMFNet & {$\omega_1=\omega_2=\omega_3=1$} & 78.969 & 90.539 & 84.207 \\
        DMFNet & Learnable $\omega_1$,$\omega_2$,$\omega_3$ & 80.12 & 90.62 & 84.54 \\
    	\hline
    \end{tabular}
\end{table}

    \begin{figure}[htbp]
	\includegraphics[width=\textwidth]{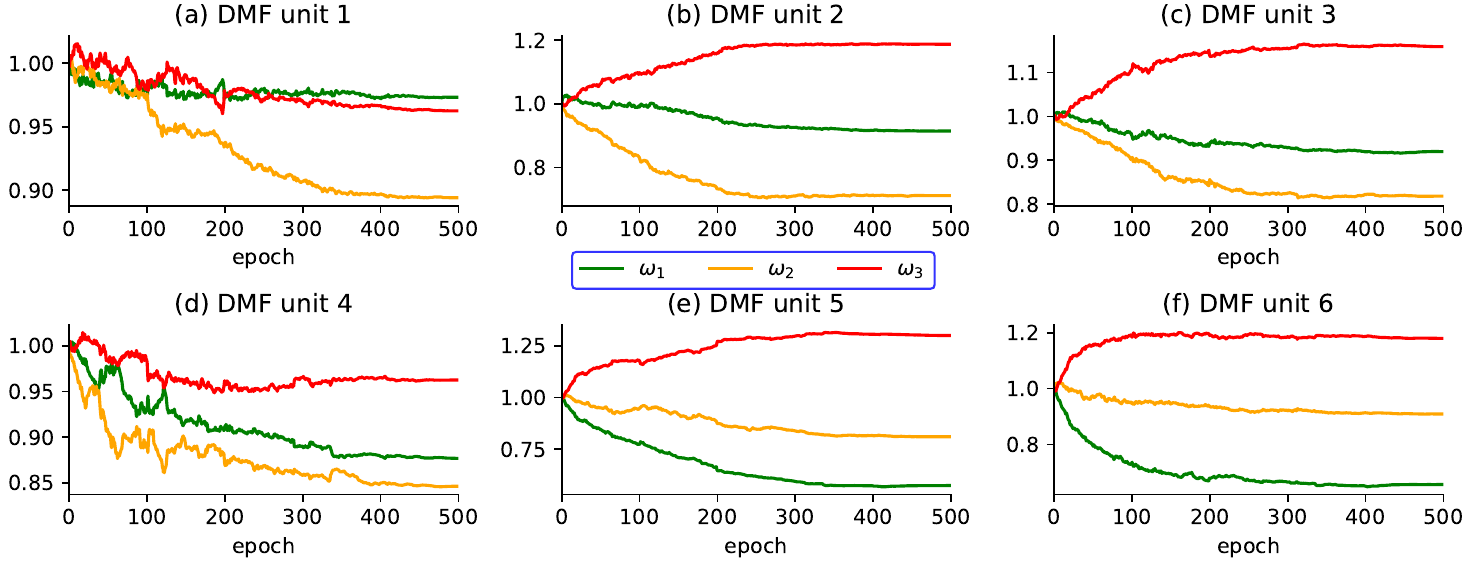}
	\caption{The changing weights $\omega_1$, $\omega_2$ and $\omega_3$ in the training process. DMF unit 1 is the $1^{st}$ DMF unit and similar to DMF units 2$\sim$6, i.e. the green blocks in \mbox{Fig. \ref{Architecture}}.}
	\label{figure3}
    \end{figure}

    The weights $\omega_1$, $\omega_2$ and $\omega_3$ in the training process are plotted in \mbox{Fig. \ref{figure3}}. It is noticed that $\omega_1$ (green line, corresponds to small receptive field) plays a major role in the first unit, and its effect is decreasing in the higher layers. While, we also observe that the network favors the dilated branch with $\omega_3$ (red line, corresponds to large receptive field), which has leading influences in DMF units 2-6. It may be because the kernel with small receptive field is not able to capture useful semantic information in the higher layers which have small dimension.

\section{Conclusion}

     In this work, we have developed a lightweight and efficient Dilated Multi-Fiber network, with only 3.88M parameters and around 27G FLOPs, that can achieve real-time inference for 3D brain tumor segmentation in MRI. To reduce the heavy computational burden in 3D convolution significantly, we explored multi-fiber units with the spirit of group convolution. Meanwhile, we introduced a learnable weighted 3D dilated convolution to gain multi-scale image representation, which is able to enhance the segmentation accuracy. The experimental results on the 2018 BraTS challenge show that our approach achieved comparable dice sores (80.12\%, 90.62\% and 84.54\% for ET, WT and TC, respectively) yet with 10$\times$ less model parameters and 50$\times$ less computational FLOPs, compared with the state-of-the-art algorithm, e.g. NVDLMED~\cite{myronenko20183d}. This makes our method more practical for handling large-scale 3D medical datasets.


\bibliographystyle{unsrt}
\bibliography{reference}

\end{document}